# Multi-input Multi-output Beta Wavelet Network: Modeling of Acoustic Units for Speech Recognition

Ridha Ejbali, Mourad Zaied and Chokri Ben Amar
REsearch Groups on Intelligent Machines, University of Sfax
Sfax, Tunisia

*Abstract*—**In this paper, we propose a novel architecture of wavelet network called Multi-input Multi-output Wavelet Network MIMOWN as a generalization of the old architecture of wavelet network. This newel prototype was applied to speech recognition application especially to model acoustic unit of speech. The originality of our work is the proposal of MIMOWN to model acoustic unit of speech. This approach was proposed to overcome limitation of old wavelet network model. The use of the multi-input multi-output architecture will allows training wavelet network on various examples of acoustic units.**

## I. INTRODUCTION

The development of robust systems for speech recognition is now one of the main issues of language processing. Many systems were based on concepts that were relatively close. However, despite those progresses, the performances of current systems remain much lower than the capacity of the human auditory system. In this respect, the prospects for improvement are strong. Economic issues related to speech recognition make this area a sector in constant evolution and pushes us to constantly imagine new approaches in order to be able to at least match or exceed the performance of our hearing.

To solve some problems of modelling and recognition of speech, we suggest a new method based on wavelet networks [21] that exploit the intrinsic properties of the speech signal, while the statistical models are concerned with statistical properties of the speech signal [2]. These models are similar to neural network for the structure and the training approach. But, training algorithms for wavelet network require a smaller number of iterations when compared with neural network.

Wavelet network model, the single-input single-output wavelet network, was introduced, firstly, by Zhang and Benveniste in 1992 [8]. This model was inspired from neural network architecture as a combination of neuronal contraption and wavelets as activation functions [20]. Those models have supplied an access to all frequency of signal thanks to the use of wavelets in the hidden layer of each neurone. It has more advantages than common networks such as faster convergence, avoiding local minimum, easy decision and adaptation of structure [4].

Despite the contribution of these models in different fields of pattern recognition [4], [5], [7], they remained limited in the field of modelling. These prototypes cannot instil entities at different occurrences. To overcome these limitations, we advanced a new model for the training of several instances of a single entity at the same time. These models are called multi-input multi-output wavelet network MIMOWN [23][24].

The idea of this architecture was presented for the first time by Zhao in [23]. The training algorithm of this model is identical to the training algorithm of the old version of wavelet networks. It is based on the forward backward algorithm. Identically to the old version, this approach takes as input a unitary random vector and according to an error defined by the characteristics of the networks, it estimate the original vector output.

Gutierrez in [25] has used the MIMOWN for the Voltammetric Electronic Tongues. In his works, the architecture is implemented with feed forward one-hidden layer architecture. In the hidden layer, he uses wavelet functions. The network training is performed using a back propagation algorithm, adjusting the connection weights along with the network parameters.

In [22], Fengqing has presented other architecture of wavelet network called multi-input single-output wavelet network. In his work, he has shown that his architecture can eliminate the useless wavelet on the network when training a data.

An improvement was made to the approach of Zhao [23] in this article. Improvement concerns the training algorithm. In this case, the input vector of the network corresponds to the vector model and not a random unitary vector. The new architecture allows training several multivalve examples.

The proposed algorithm car train a MIMOWN on original vector and not on unitary random vector as in [23][24][25].

Thinks to those models, the training system can model a variety of occurrences of a single entity by a single acoustic wavelet network [22]. The newel approach can be seen as a superposition of finite number of single-input single-output wavelet network.

Those prototypes are used to model acoustic unit of speech to be used on speech recognition system.

In this paper, we propose a newel approach to model acoustic units of speech for the recognition of isolated words based on MIMOWN. These new models are similar to multilayer neural network for the structure and the training approach. Compared to old wavelet network structure, those prototypes can represent several acoustic units by one network.

Our systems, based on wavelet network have adaptable parameters; dilation, translation, and weights. Initial values of weights are made randomly. Then, parameters are optimized automatically during training [6]. To update parameters, gradient method has been applied by using momentum. Quadratic cost function is used for error minimization.

This paper includes four parts. In the first part, we present the techniques used to construct the newel prototype MIMOWN. In the second part, we detail the training approach of MIMOWN. The third part presents the recognition part of our system. The finale part illustrates the results that crown our approach.

## II. Technical Background

### A. *Beta wavlet*

According to works published in [17], [18] and [19] Beta function is a wavelet and from it we can build a network called Beta wavelet.

It satisfies the properties of a wavelet namely:

- ✓ Admissibility : $0 < C_\psi = 2\pi \int \frac{\|\hat{\psi}(w)\|^2}{\|w\|} dw < \infty$
- ✓ Localization: wavelet is a function of $L^2(IR)$ with the property of location; it decreases rapidly on both sides of its domain.
- ✓ Oscillation : wavelet is a function of $L^2(IR)$ , integrated and oscillating enough to have zero as integral: $\int \psi(t)dt = 0 \Leftrightarrow TF(\psi(0)) = 0$
- ✓ The translation and dilation : The wavelet analysis involves a family of copies of itself, translates and dilated: $\psi_{ab}(t) = \frac{1}{\sqrt{a}}\psi(\frac{t-b}{a})$ *with* $a,b \in IR, a > 0$

### B. *Wavelet Network*

The field of wavelet network is new, although some attempts have recently been conducted to build a theoretical basis and several applications in various fields. The use of wavelet network began with the use of Gabor wavelets in classification and image recognition [1], [3], [8].

As shown in the figure below, wavelet network can be seen as a neural network with wavelet in the hidden layers [9], [10], [11]:

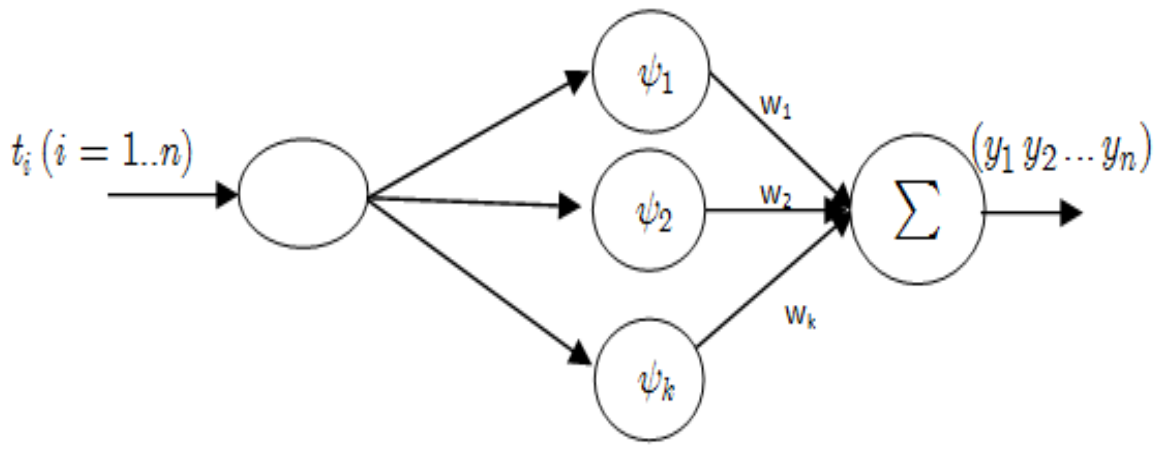

Figure 1. Example of a figure caption.

The reconstruction formula of a signal after processing by a continuous wavelet is given by:

$$f(x) = \frac{1}{C_\psi}\int_{IR_+}\int_{IR} W_f(a,b)\frac{1}{\sqrt{a}}\psi\left(\frac{x-b}{a}\right)da\,db \quad (1)$$

This formula gives the expression of the signal $f$ in the form of an integral over all possible translations and dilations of a mother wavelet $\psi$ . Suppose we have only a finite number $N_w$ wavelet $\psi_j$ obtained from the mother wavelet $\psi$ . We can then reformulate the equation (1) as following:

$$f(x) \approx \sum_{j=1}^{N_w} w_j \psi_j(x) \quad (2)$$

We can consider the expression of the equation (2) as a decomposition of the function $f$ to a sum of weights $w_i$ and wavelets $\psi_i$ .

To define a wavelet network, we start with taking a family of $N_w$ wavelets $\psi = \{\psi_1 ... \psi_{N_w}\}$ with different parameters of dilatation and translation that can be chosen randomly at this point.

The old architecture of wavelet networks can learn more examples but this example must be mono-valued. That's why we thought to extend this architecture for new models of wavelet network that can learn several examples which are multi-valued.

### C. *Idea of multi-input multi-ouput wavelet network*

*1)* Architetcure: The old version of wavelet network has certain proximity of the multilayer perceptron network MLP architecture with one hidden layer. The essential difference between these two networks is the nature of transfer functions used by the hidden cells.

The new proposed architecture is a wavelet network with multi-input multi-output shown in figure 2. This network is seen as the superposition of a finite set of wavelet network with one input and one output.

*2)* Limits: The superposition of three patterns (a), (b) and (c) constitute the new MIMOWN. This new architecture will, unlike the old, take as input the vector of learning and not an impulse vector. In addition to the shift from a single-input single-output architecture provides a new model to multiple outputs.

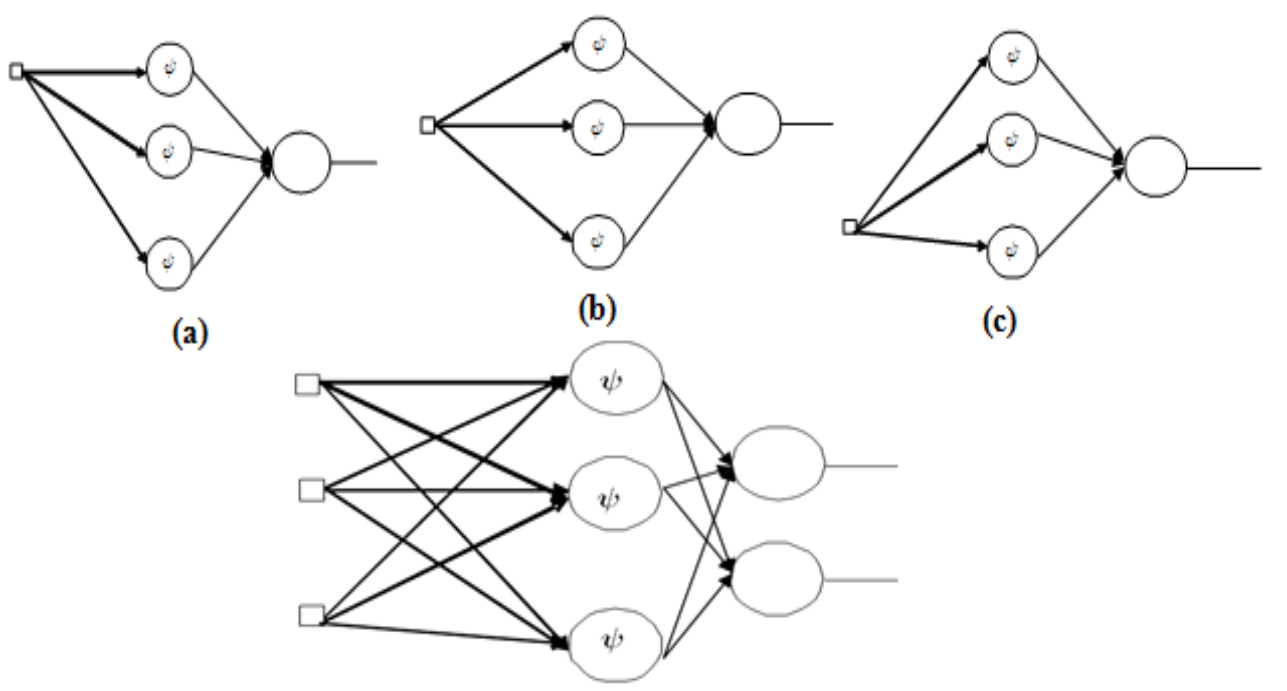

Figure 2. Building of multi-input multi-output network of wavelet.

*D. Trainig of a milti-input multi-outout wavelet network*

*1) Introduction:* In this section, we will present Zhao architecture [23] and our architecture considered as generalization of the training algorithm of Zhang's networks to train MIMOWN.

*2) Approach proposed by Zhao to train multi-input multi-output wavelet network:* In his work, Zhao has changed the functions of activations of a neural network with wavelet. For training architecture, Zhao uses the method of wavelet network learning single-input single-output.

The following figure illustrates the Zhao wavelet network architecture:

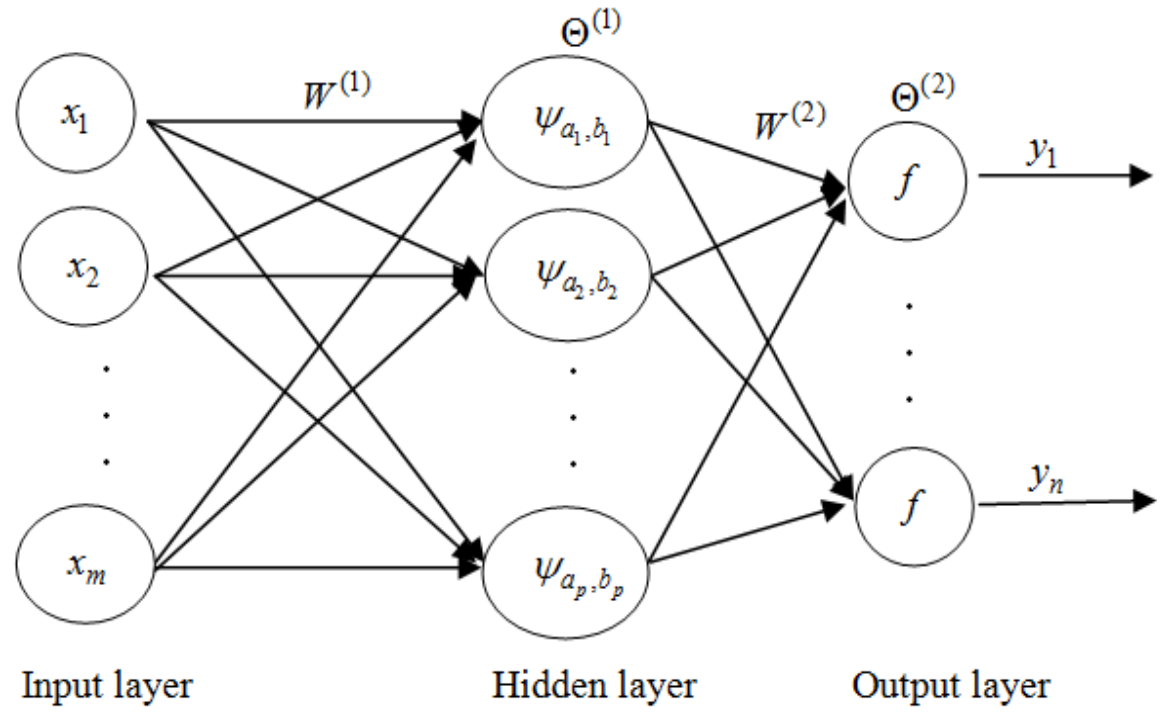

Figure 3. Zaho architecture of wavelet network.

The training phase starts with the preparation of unitary random vectors of size equal to the base vector. These vectors constitute the input of the network. The calculation of the weights of networks is based on a calculated distance between the output vector of the network and the original vector of training. The error between these two vectors is determined by the characteristics of the network such as the number of wavelets in the hidden layer of the network. If the error set is reached, the learning process stops otherwise it will change the parameters of the wavelets in the hidden layer to improve the value of the output vector.

*3) Mathematical concepts:*

✓ *Periodic Function*

A function $f$ is periodic if there is a strictly positive real $T$ such that $\forall x \in D_f$, we have

$$x+T \in D_f \text{ and } f(x+T) = f(x) \quad (3)$$

We call the period of the function $f$ the smallest real $T$ verifying the property **(3)**. If $f$ is a function of period $T$, then $\forall x \in D_f$, we have $f(x+kT) = f(x)$ with $k \in IN$.

The study interval of a periodic function can be reduced to an interval covering a single period.

✓ *Taylor formula for polynomial functions*

$P$ is a polynomial function, if degP $=$ n with $n \in IN^*$. Using the Taylor formula for the polynomial functions, we have

$$P(x) = \sum_{k=0}^{n} \frac{P^{(k)}(x_0)}{k!}(x-x_0)^k = P(x_0) + P'(x_0)(x-x_0) + P''(x_0)\frac{(x-x_0)^2}{2} + .... + P^{(n)}(x_0)\frac{(x-x_0)^n}{n!} \quad (4)$$

If $p > 0$ and $q > 0$

$$\beta(x) = \begin{cases} \left(\dfrac{x-x_0}{x_c-x_0}\right)^p \left(\dfrac{x_1-x}{x_1-x_c}\right)^q & \text{if } x \in ]x_0, x_1[ \text{ with } x_c = \dfrac{px_1 - x_0}{p+q} \\ 0 & \text{else} \end{cases} \quad (5)$$

Applying (4) for the Beta function (polynomial for p and q integers)[16]

$$\begin{aligned} \beta(x) &= \sum_{k=0}^{n} \frac{\beta^{(k)}(x_0)}{k!}(x-x_0)^k \text{ with } n=p+q \\ &= \sum_{k=0}^{n} \frac{Q_k(x_0)\beta(x_0)}{k!}(x-x_0)^k \quad \text{using (5)} \\ &= \left(\sum_{k=0}^{n} \frac{Q_k(x_0)}{k!}(x-x_0)^k\right)\beta(x_0) \end{aligned} \quad (6)$$

*4) General idea of the algorithm: As shown in figure 4, it is a multi-layer wavelet network.*

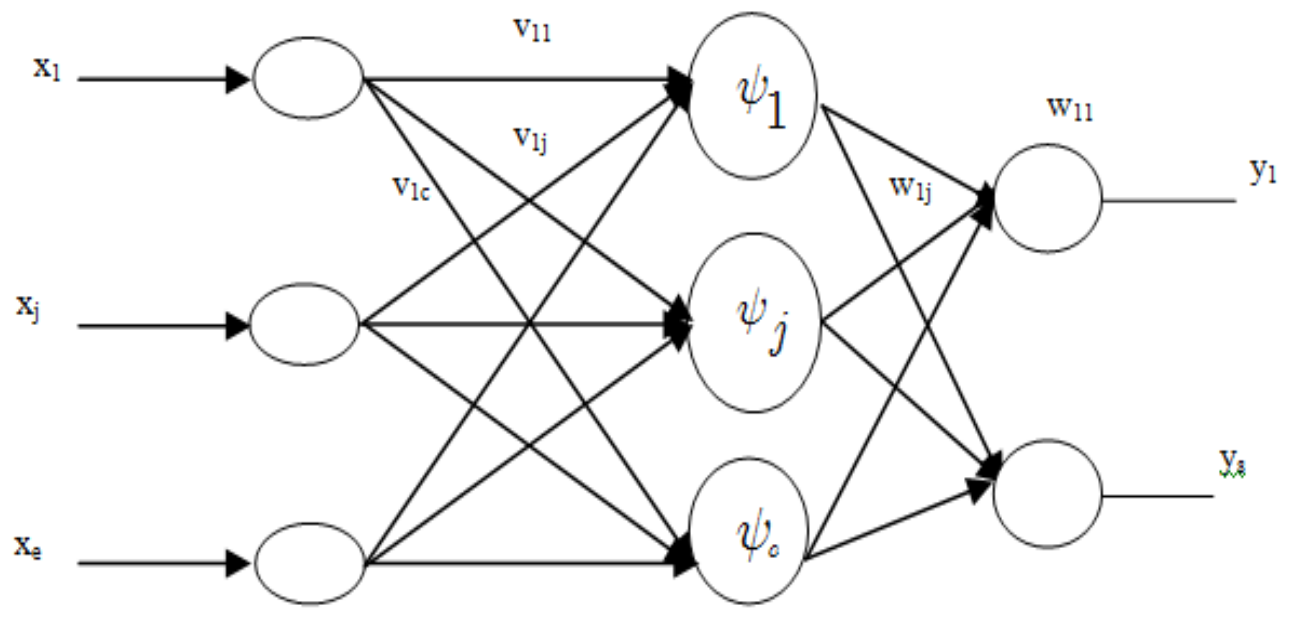

Figure 4. Multi-input multi-ouput wavelet network.

The network presented in the previous figure is constructed by a superposition of networks Zhang. For signal: $Y = (y_1, y_2, ......., y_s)$ The network of Zhang expressed as follows:

$$(I)\begin{cases} y_1 = \alpha_1\psi_1(A_1) + \alpha_2\psi_2(A_1) + ..... + \alpha_c\psi_c(A_1) \\ ........................................................ \\ y_s = \alpha_1\psi_1(A_s) + \alpha_2\psi_2(A_s) + ..... + \alpha_c\psi_c(A_s) \end{cases} \quad (7)$$

While the MIMOWN expressed by:

$$(II)\begin{cases} y_1 = w_{11}\psi_1(A_1) + w_{12}\psi_2(A_2) + ..... + w_{1c}\psi_c(A_c) \\ ............................................................ \\ y_s = w_{s1}\psi_1(A_1) + w_{s2}\psi_2(A_2) + ..... + w_{sc}\psi_c(A_c) \end{cases} \quad (8)$$

With $A_i$ the entry of the neuron number $i$, $\alpha_i$ represent the weights of the connections of the network of Zhang and $w_{ij}$ are the weights of connections of the MIMOWN.

In each scale, we can use a single wavelet instead of $2^{j-m}$ because:

On the scale $m$ in the library, there are $2^{j-m}$ wavelets $(\psi_i)_{1\le i\le 2^{j-m}}$. The graphical representations of wavelet scales are adjacent and have the same size. We suppose that they form a single function, so the latter is periodic with period T. The size, the support, of wavelet is $T = \frac{N}{2^{j-m}}$ with $j = \log_2(N)$ is the number of scales to cover the whole signal. $N$ is the size of the signal.

We have $\forall\ 1 \le i \le j \le 2^{j-m}$ for $x''-x' = (j-i)T, \psi_i(x') = \psi_j(x'')$

For a signal of size N = 256, we can analyze up to the scale N = 8 (j = 8). In the 6th scale, there are 4 wavelets.

## III. TRAINING PROCESS

Let $X = (x_1, x_2, ..., x_e)$ the input vector, $\psi = (\psi_1, \psi_2, ..., \psi_c)$ the vector containing the wavelet to be used in the hidden layer and $Y = (y_1, y_2, ..., y_s)$ the output vector.

- ✓ For the weights connecting the input layer and hidden layer $(V_{ij})_{\substack{1\le i\le e \\ 1\le j\le c}}$

We calculated $V_{i,j}$ in order to have as input of the j<sup></sup>ème wavelet $A_j = j$.

Hence $V_{i,j} = \frac{j}{\text{length of the signal}*X(i)}$

- ✓ For the weights connecting the hidden layer and the input layer $(W_{ij})_{\substack{1\le i\le c \\ 1\le j\le s}}$

Let $g$ be a signal, $f_i$ is a family of functions that form a basic family of dual functions $\tilde{f}_i$ then there exist $\alpha_i$ weights such as:

$$g = \sum_i \alpha_i f_i \quad (9)$$

Since the wavelet used is Beta (is a polynomial for p and q integers) using Taylor's theorem, from equation **(9)**, we can write $\psi_i(A_j) = c_{ij}\psi_i(A_i)$ with $c_{ij} = \sum_{i=1}^{p+q} \frac{Q_k(A_i)}{k!}(A_j - A_i)^k$.

$$(I')\begin{cases} y_1 = \alpha_1 c_{11}\psi_1(A_1) + \alpha_2 c_{12}\psi_2(A_2) + ..... + \alpha_c c_{1c}\psi_c(A_c) \\ ............................................................ \\ y_s = \alpha_1 c_{s1}\psi_1(A_1) + \alpha_2 c_{s2}\psi_2(A_2) + ..... + \alpha_c c_{sc}\psi_c(A_c) \end{cases} \quad (10)$$

And by identifying with the system (II) we have: $w_{ij} = \alpha_i c_{ij}$

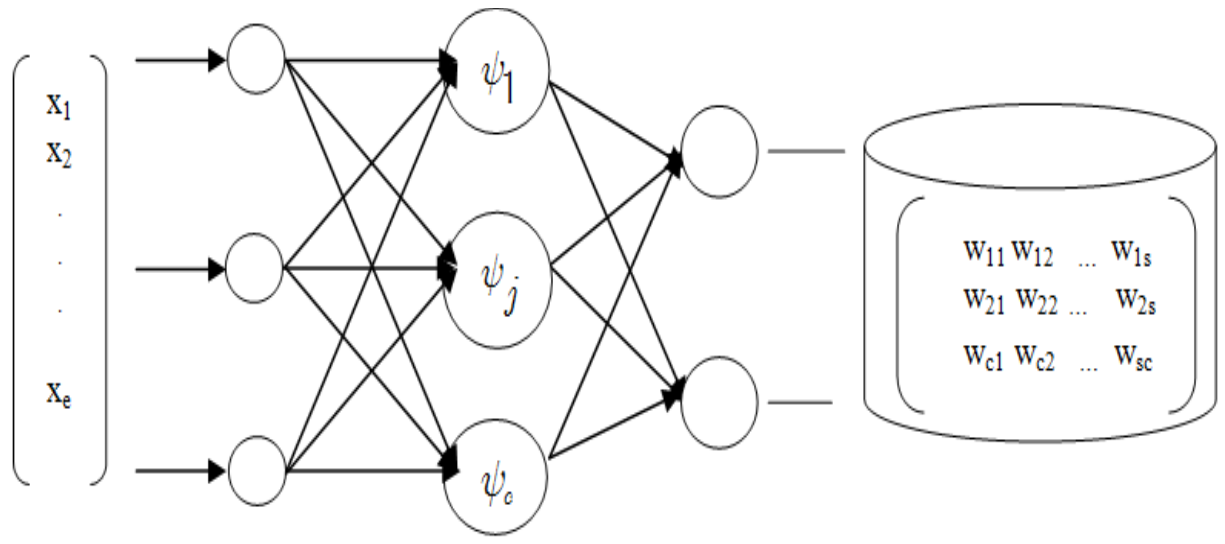

Figure 5. Training phase.

The performance of approximation networks is the basis of parameters estimated by measuring the mean squared error, expressed by the following formula:

$$EQM = \frac{1}{M*N}\sum_{i=1}^{N}\sum_{j=1}^{M}(A(i,j) - B(i,j))^2 \quad (11)$$

Such as A and B represent the coefficients of input and output network, while M and N dimensions.

## IV. RECOGNITION PROCESS

During the recognition phase, the system will construct a new vector for each element of the test. Every element of the test will be the entry of all networks-based learning. The system will modify the weights of the network learning approach to maximize the input test vector.

At the end of the phase, we will save the weight from each network after approximating the test vector.

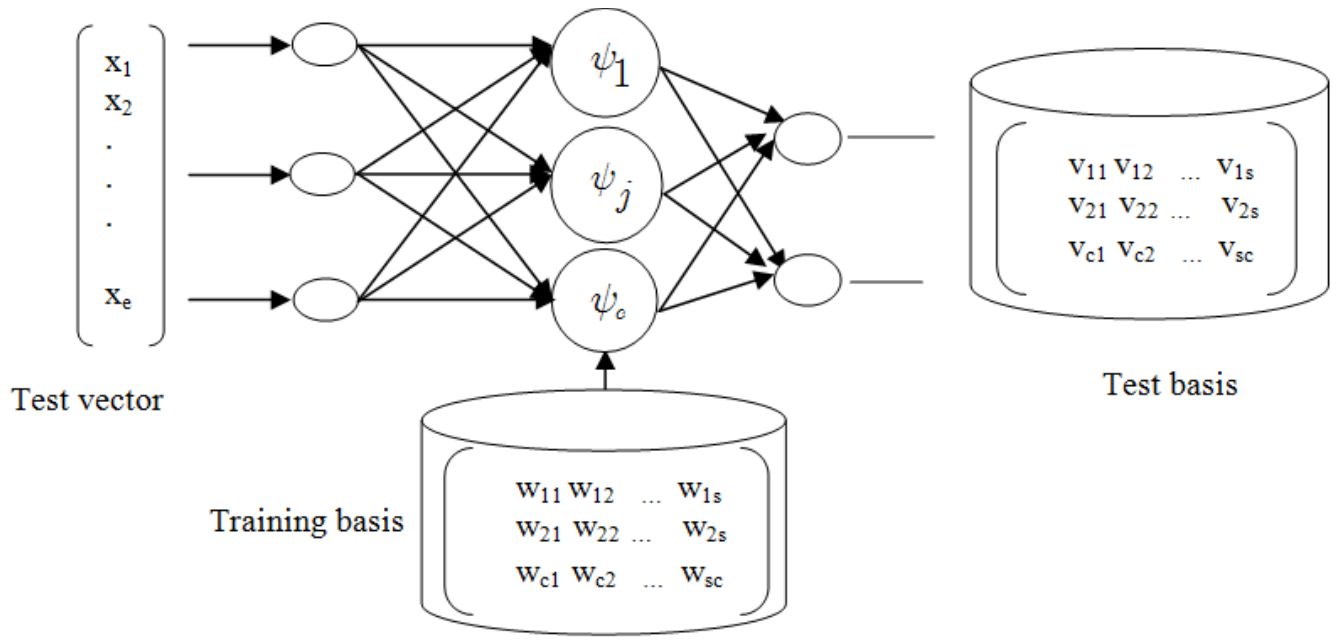

Figure 6. Reocgnition phase.

## V. Corpus

We have tested our modelling approach on two corpuses:

The first corpus was recorded by 12 speakers (6 women and 6 men) from works of (Boudraa and Boudraa, 1998) [12]. We segmented this corpus manually by PRAAT to Arabic words and we have chosen 19 different words. We have added sound effects for each word to test robustness of our approach. Finally, we got 912 words. The second corpus was recorded by 14 speakers. It was about Tunisian city name (24 cities). We have added the same sound effects to each word. Finally, we got 1344 words.

We have chosen Mel-Frequency Cepstral Coefficients MFCC and Perceptual Linear Predictive (PLP) coefficients to represent acoustic data. Our choice of acoustic representation is based on the specificities of these types.

- ✓ MFCC coefficients contain features around Fast Fourier Transform (FFT) and Discrete Cosine Transform (DCT) converted on Mel scale. This method used mostly to represent a signal in speech recognition, because of its robustness. Its advantage is that the coefficients are uncorrelated.
- ✓ PLP technique aims at estimating the parameters of an autoregressive filter and all-pole modelling to better auditory spectrum. It takes into account many limitations and characteristics of human speech production and hearing to reduce a number of direct waveform samples into a few numbers that represent the perceived frequency concentrations and widths.

## VI. Tools

We used PRAAT to record the speech and to handle signals and words segmentation. Recognition system of Arabic word using HMM has been implemented with the HTK library version 4.0 [13]. Recognition system of Arabic word using WN was implemented using MATLAB.

## VII. Results

To test the new architecture of modelling of acoustic units of speech, we have implemented a recognition system of Arabic word [4]. The results of this approach, MIMOWN, of modelling was compared to results given by a recognition system of Arabic words based on Hidden Markov Models HMM [14], [15] and the old architecture of Wavelet Network WN. The following figures illustrate the recognition rate of the three modelling approaches using the same type and number of MFCC coefficient describing units.

Figure 7 shows the recognition rate of the three modeling approaches applied to the corpus of Tunisian cities names on MFCC coefficients.

According to Figure 7, we deduce that our new approach based on MIMOWN has produced good results compared to WN and HMM based system.

The recognition rates of the same systems are also compared using PLP coefficients. Those results are illustrated on figure 8.

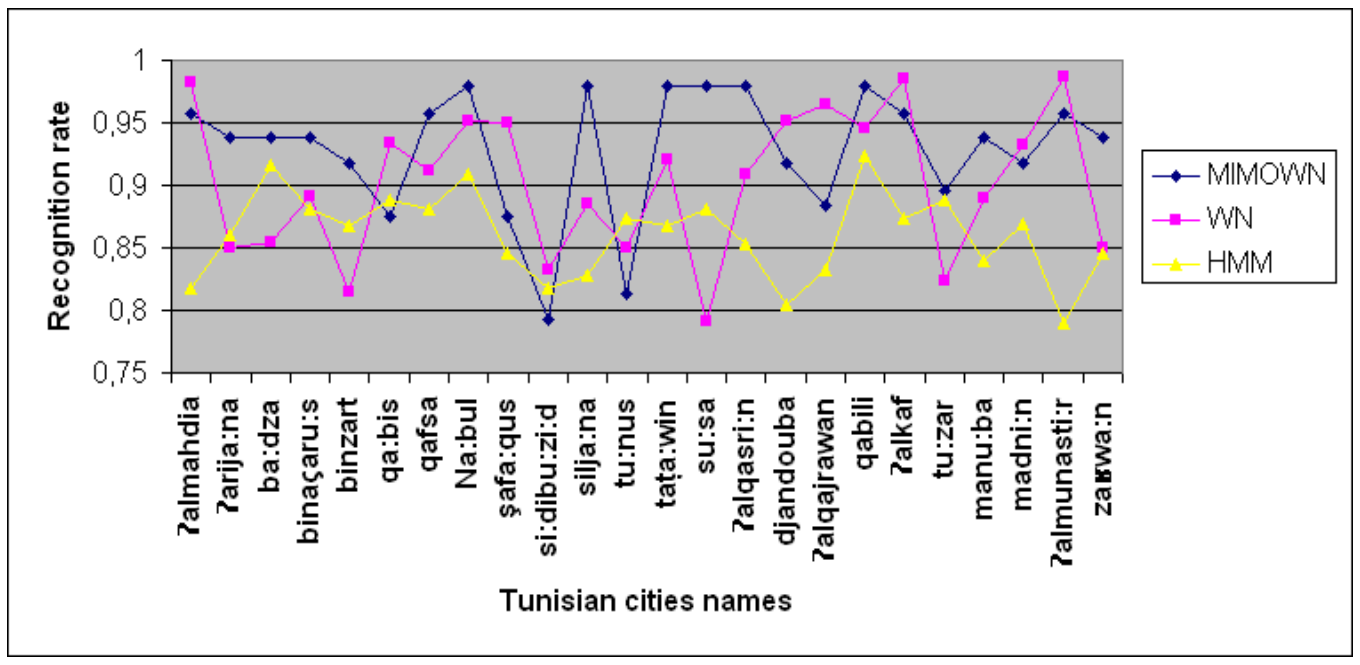

Figure 7. Recognition rates of the Tunisian cities names

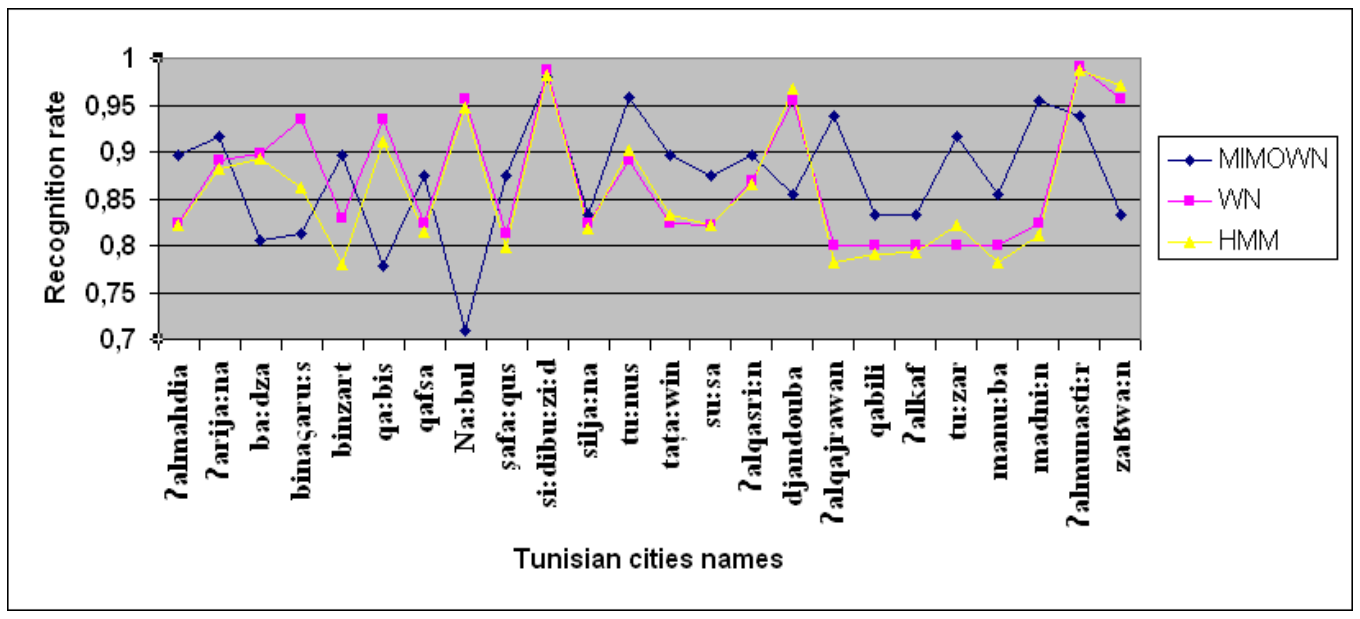

Figure 8. Recognition rates of the Tunisian cities names

According to figure 8, we can say that our approach gives good results comparing to the old wavelet network and HMM approach.

Figure 9 shows the recognition rate of the three modelling approaches applied to the corpus of Arabic words on MFCC coefficients.

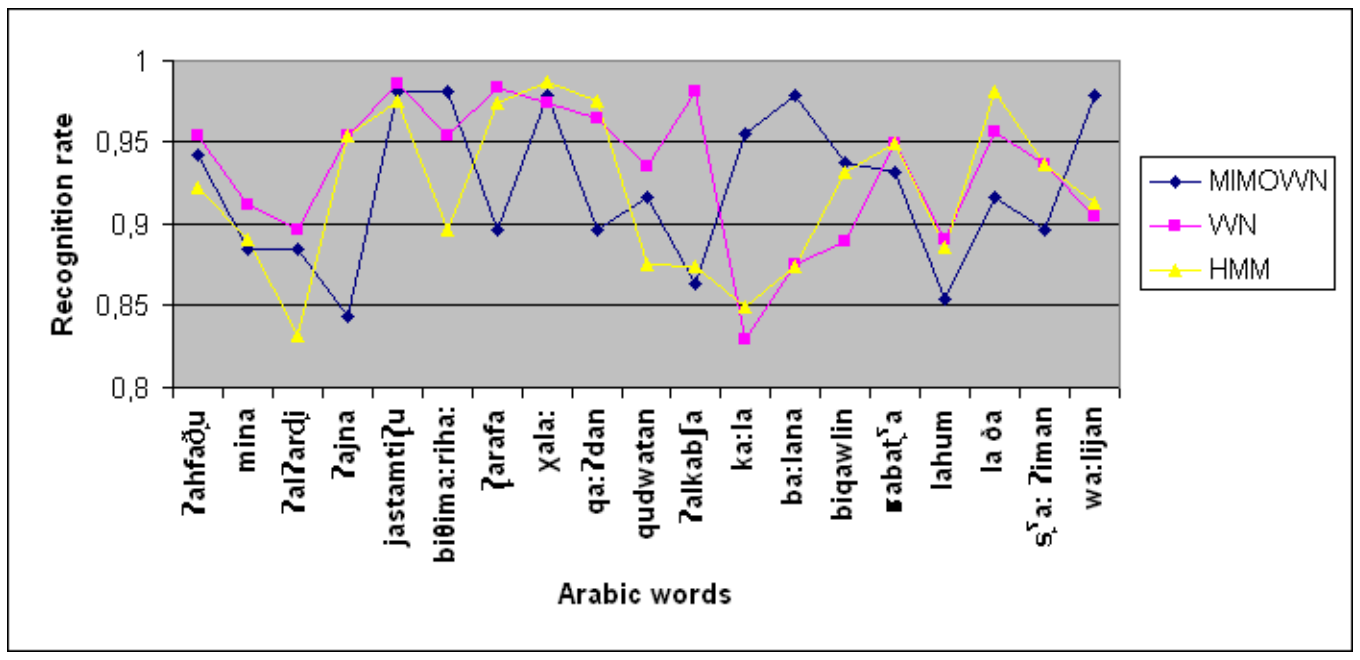

Figure 9. Recognition Rates of Arabic words

According to figure 9, the results given by multi-input multi-output approach are comparable to the wavelet network and hidden Markov model approach.

Increasing the number of training data over networks will refine multi-input multi-output. Each new sound unit will add a set of nodes in the hidden layer networks thus a better identification of test units.

Figure 10 illustrate recognition rate of the tree systems on PLP coefficients.

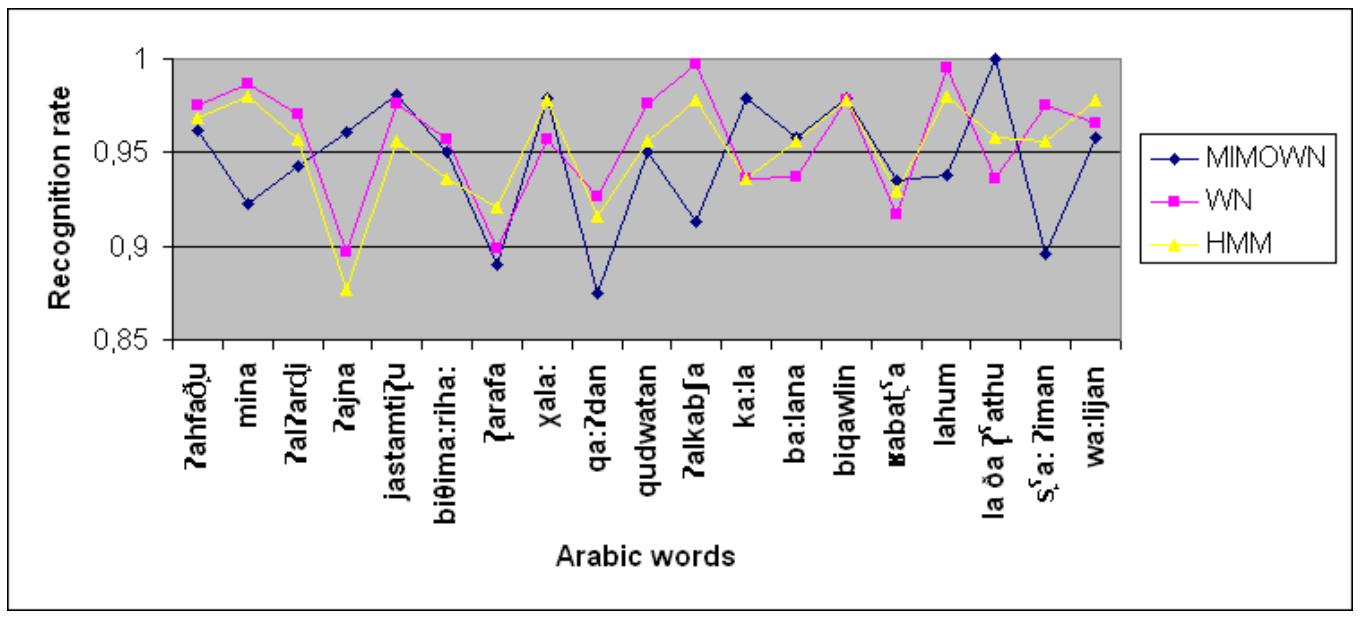

Figure 10. Recognition Rates of Arabic words

Based on the previous figures, we can realize that our approach based on MIMOWN gave better results compared to the approach based on WN and HMM.

These results can be improved by increasing the size of the training base, getting a better quality of recordings, choosing other types of wavelet…etc.

This new architecture can learn several examples with multiple values. It will allow the modeling of instances of an acoustic unit by a wavelet network. Thanks to this architecture, each unit will be represented by a matrix of weight. The updating of the training base will not require the creation of a new network for each new entry, but the update of the associated network. In case the new unit does not exist in the database, it is evident to create a new model that is network wavelet.

## VIII. Conclusion

Given the success of the old architecture of wavelet networks, single-input single-output architecture, it was interesting to study the possibility of generalizing this prototype to remedy the gaps of the old version. This work focuses on suggesting a new architecture of wavelet networks, MIMOWN. This new architecture has been proposed for modelling the acoustic units of speech. This new prototype will allow the modelling of several examples. It takes as input the coefficients of each unit and not a vector pulse.

Tests made on cities words corpus and Arabic word corpus show that our approach based on MIMOWN gives good results compared to WN and HMM model based system.

Interpreting the findings of this new modelling technique, we came to the conclusion that its adoption in large vocabulary applications and real application will improve performance of the speech recognition system

## Acknowledgment

The authors would like to acknowledge the financial support of this work by grants from General Direction of Scientific Research (DGRST), Tunisia, under the ARUB program.

## Authors Profile

Ridha Ejbali is a PhD student in the REsearch Group on Intelligent Machines, National Engineering School of Sfax, university of Sfax Tunisia. He received the Master degree of science from the National Engineering School of Sfax (ENIS) in 2006. He obtained the degree of Computer Engineer from the National Engineering School of Sfax (ENIS) in 2004. He was assistant technologist at the Higher Institute of Technological Studies, Kebili Tunisia since 2005. He joined the faculty of sciences of Gabes (FSG) where he is an assistant in the Department computer sciences since 2012. His research area is now in pattern recognition and machine learning using Wavelets and Wavelet networks theories. He is IEEE Graduate Student Member, SPS society and Assistant in the Faculty of Sciences of Gabes Tunisia.

Dr. Mourad ZAIED received the Ph.D degree in Computer Engineering and the Master of science (DEA : Diploma in Higher Applied Studies) from the National Engineering School of Sfax (ENIS) respectively in 2008 and in 2003. He obtained the degree of Computer Engineer from the National Engineering School of Monastir (ENIM) in 1995. Since 1997 he served in several institutes and faculties in the university of Gabes as a teaching assistant. He joined the National Engineering School of Gabes (ENIG) where he is an assistant professor in the Department of Electrical Engineering since 2007. His research area is now in pattern recognition and machine learning using Wavelets and Wavelet networks theories. He is an IEEE member and SMC society member.

Prof. Chokri BEN AMAR received the B.S. degree in Electrical Engineering from the National Engineering School of Sfax (ENIS) in 1989, the M.S. and PhD degrees in Computer Engineering from the National Institute of Applied Sciences in Lyon, France, in 1990 and 1994, respectively. He spent one year at the University of "Haute Savoie" (France) as a teaching assistant and researcher before joining the higher School of Sciences and Techniques of Tunis as Assistant Professor in 1995. In 1999, he joined the Sfax University (USS), where he is currently an associate professor in the Department of Electrical Engineering of the National Engineering School of Sfax (ENIS), and the Vice director of the Research Group on Intelligent Machines (REGIM). His research interests include Computer Vision and Image and video analysis. These research activities are centered on Wavelets and Wavelet networks and their applications to data Classification and approximation, Pattern Recognition and image and video coding, indexing and watermarking. He is a senior member of IEEE, and the chair of the IEEE SPS Tunisia Chapter sinc 2009.